\documentclass{lcs2}

\usepackage{multicol, multirow}
\usepackage{caption}   % provides \captionof
\usepackage{array}
\usepackage{arydshln}

\papergithub{https://github.com/parmanu-lcs2}  %optional
\paperwebsite{https://parmanu.lcs2.in}  %optional
\newcommand{\pruner}{\textsc{Maestro}}
\papertitle{It Takes a \pruner\ To Prune Bad Experts} %change this

\papershortauthors{P. Goel, A. Maheshwari and T. Chakraborty} %change this
\paperauthors{%
  Palaash Goel \textsuperscript{\tiny 1} \quad
  Ayush Maheshwari \textsuperscript{\tiny 3} \quad
  Tanmoy Chakraborty \textsuperscript{\tiny {1,2}}%
} %change this

\paperaffil{\textsuperscript{1} Yardi School of Artificial Intelligence, Indian Institute of Technology Delhi\\
\textsuperscript{2} Department of Electrical Engineering, Indian Institute of Technology Delhi\\
\textsuperscript{3} NVIDIA, India} %change this
\paperemails{{goelpalaash@scai.iitd.ac.in, aymaheshwari@nvidia.com, tanchak@iitd.ac.in}} %change this

\paperkeywords{LLM compression, structured pruning, mixture-of-experts, LLM efficiency} %optional %change this

\paperabstract{Sparsely-activated Mixture-of-Experts (MoE) language models achieve remarkable inference efficiency by activating only a small fraction of parameters per token, yet their full expert banks reside in memory at all times, creating a prohibitive deployment bottleneck. Existing structured pruning methods, largely designed for dense transformers, assess expert importance using locally derived heuristics that are blind to the interdependent nature of MoE routing. We introduce \pruner\ (\textbf{M}arkov-chain \textbf{A}pproximated \textbf{E}xpert \textbf{S}parsification via \textbf{T}ransition-based \textbf{RO}uting), a structured pruning framework designed for MoE architectures that models autoregressive expert activation trajectories as Ergodic Markov chains whose stationary distributions encode cross-layer dependencies, yielding a globally aware importance heuristic. 
Evaluated across five diverse domains including Safety, Bias, and Ethics, \pruner\ outperforms state-of-the-art baselines by up to 10.61\% in average performance retention under a strict 50\% compression regime, while exhibiting substantially lower cross-task variance, indicating that global, routing-congruent pruning produces models that generalize more consistently across heterogeneous tasks.
}

\begin{document}

\makelabtitle

%%%%%%%%%%%%%%%%%%%%%%%%%%%%%%%%%%%%%%%%%%%%%%%%%
\section{Introduction}
\label{sec:introduction}

Sparsely-activated Mixture-of-Experts (MoE) architectures have emerged as the dominant recipe for scaling large language models without a commensurate increase in inference compute \citep{jiang2024mixtralexperts, deepseekai2025deepseekr1incentivizingreasoningcapability, yang2025qwen3technicalreport, openai2025gptoss120bgptoss20bmodel, nvidia2025nvidianemotron3efficient}. By replacing the dense feed-forward block of each transformer layer with a bank of $E$ expert sub-networks and a learned router that dispatches each token to only $k \ll E$ of them, MoE models decouple parameter count from per-token FLOPs. Such models routinely instantiate tens to hundreds of billions of parameters while activating only a few billion per token, yielding inference throughput comparable to far smaller dense models.

This same design choice, however, is the source of a sharply contradictory deployment profile. While MoE models are \textit{computationally} efficient at inference time, they remain \textit{spatially} prohibitive: the full bank of experts must reside in memory, regardless of how few are selected for any given token. 
% A model that activates the equivalent of $3$B parameters per token may nevertheless require the storage and high-bandwidth memory of a $30$B or $120$B dense checkpoint. 
This results in a deployment bottleneck wherein state-of-the-art MoE models do not fit on commodity accelerators, are expensive to host even on data-centre hardware, and are effectively excluded from on-device, edge, and memory-constrained settings where their per-token compute profile would otherwise be ideal. Closing the gap between activated and total parameters is therefore a central practical problem for the deployment of modern MoE LLMs.

A natural response is structured pruning: permanently removing parameters from the model so that both compute and memory drop simultaneously. The pruning literature on large language models (LLMs), however, has overwhelmingly targeted \textit{dense} transformers \citep{ma2023llmpruner, ashkboos2024slicegptcompresslargelanguage, men-etal-2025-shortgpt, guo2025slimllmaccuratestructuredpruning, shopkhoev2026replacemenetworksimplificationdepth, song2024sleb, prunenet}. Methods such as magnitude- \citep{magnitude-pruning}, activation- \citep{wanda}, and gradient-based pruning \citep{ma2023llmpruner, optimalbraincompression}, layer dropping \citep{shopkhoev2026replacemenetworksimplificationdepth, song2024sleb, men-etal-2025-shortgpt}, and width reduction \citep{ma2023llmpruner, ashkboos2024slicegptcompresslargelanguage, prunenet} are designed around the assumption that every parameter participates in every forward pass and can therefore be scored by global, token-averaged signals. These assumptions are blind to the routing behavior and long-range inter-expert dependencies that exist within MoE models and thus, do not transfer cleanly to them. 
% An expert that is rarely selected by the router contributes almost nothing to the model's outputs, yet occupies the same memory as a heavily-used expert; conversely, an expert whose weights have small magnitude may still be routed to frequently and carry information that no other expert replicates. The router, which is the very mechanism that determines which parameters are alive on any given token, is invisible to dense pruning criteria, and yet it encodes precisely the signal a principled MoE pruner ought to consult.
Additionally, prior MoE-centric efforts tend to rely on local, layer-isolated heuristics such as marginal performance degradation upon the removal of a subset of experts \citep{lu-etal-2024-expert-sparsity} or router-weighted activations produced at each layer \citep{xie2024moeprunerpruningmixtureofexpertslarge}.
% Additionally, while works such as \citet{yang2025moepathfindertrajectorydrivenexpert} attempt to ground pruning decisions globally by planning over cross-layer paths, the weights assigned to those paths are products of independent local marginals — meaning global optimality considerations are made at the level of aggregation, not at the level of the signal being aggregated due to which no true inter-layer conditional dependencies are ever captured. 
Such criteria treat each MoE layer and/or expert independently and ignore the fact that routing decisions are correlated across layers. 
An expert that is moderately used at layer $\ell$ may, in practice, only be reached by tokens that subsequently visit a rarely-used expert at layer $\ell{+}1$; its true contribution to the model's behavior is a property of the joint distribution over routing trajectories, not of any single layer's marginal. A principled importance signal for MoE pruning should therefore capture how routing flows through the entire stack of expert banks, not merely how often each expert is hit in isolation.

In this work, we introduce such a signal in the form of \pruner, a framework for \textbf{M}arkov-chain \textbf{A}pproximated \textbf{E}xpert \textbf{S}parsification via \textbf{T}ransition-based \textbf{RO}uting. We model the sequence of $(\texttt{layer},\texttt{expert})$ pairs visited by a token as a Markov chain over all expert slots of an MoE model, with transitions estimated empirically over a small calibration corpus in an autoregressive manner. This yields a single, time-homogeneous kernel whose stationary distribution reflects global inter-expert dependencies and assigns each expert slot a long-run mass: the fraction of routing time the model spends in that slot under its own generation distribution. Experts with the smallest stationary mass are, by construction, the least frequently visited across all routing trajectories and are the natural candidates for removal. 
% Pruning is then performed structurally by physically slicing expert and router parameter tensors in order to reduce parameter counts. 
% A short LoRA recovery fine-tuning stage on attention projections, with experts and routers held frozen, restores task performance without revisiting the structural decisions made by the chain.

By leveraging global routing trajectories that arise in MoE models, our proposed framework avoids making locally myopic pruning decisions and instead grounds expert importance in the model's own autoregressive generation behaviour, capturing joint dependencies across the entire architecture that per-layer heuristics structurally cannot.
 
% The proposed criterion is purely a property of routing, and the implementation is purely a property of MoE primitives (a router producing top-$k$ indices and an expert bank applying expert-indexed feed-forwards). 
% We therefore apply identical machinery to three contemporary MoE families---GPT-OSS, Qwen3.5-MoE, and Llama-4---with only superficial, attribute-name-level differences.

\textbf{Contributions.} Our contributions are as follows.
\begin{itemize}[noitemsep, topsep=0pt, leftmargin=*]
    \item We propose a novel pruning framework, \pruner, that formalizes expert-level routing in MoE LLMs as a cyclic Markov chain over states of the form $(\texttt{layer},\texttt{expert})$.
    % and propose its stationary distribution as a principled and theoretically grounded MoE-aware heuristic for structural pruning. 
    % \item To the best of our knowledge, our proposed expert-level importance heuristic is the first to be congruent with global routing decisions, thus yielding a pruning mechanism that directly incorporates cross-layer dependencies instead of computing pruned configurations from signals that remain locally derived.
    \item We employ the chain's stationary distribution as a theoretically-grounded expert importance heuristic that is congruent with global routing decisions, thus yielding a pruning mechanism that directly incorporates cross-layer dependencies instead of computing pruned configurations from locally derived signals.
    % \item We evaluate \pruner\ on a comprehensive set of 17 benchmarks spanning five diverse domains, including Safety, Bias, and Ethics, across two contemporary MoE families --- GPT-OSS and Qwen 3. 
    \item We demonstrate that \pruner\ consistently outperforms current state-of-the-art baselines by up to 10.61\% on average performance retention even under a strict 50\% compression regime, while demonstrating notably low per-task standard deviation. Our evaluations are conducted on a comprehensive set of 17 benchmarks spanning five diverse domains, including Safety, Bias, and Ethics, across two contemporary MoE families: GPT-OSS and Qwen3.
    % show that it consistently outperforms current state-of-the-art baselines across a comprehensive set of 18 benchmarks spanning 5 different domains, including Safety, Bias, and Ethics.
\end{itemize}

\section{Related Works}
The computational demands of LLMs have grown substantially due to architectural over-parameterization \citep{frankle2019lotterytickethypothesisfinding, michel2019sixteenheadsreallybetter, ma2023llmpruner, k-etal-2025-llm-overparam}. While techniques such as quantization \citep{bhandare2019efficient8bitquantizationtransformer, zeroquant, frantar2023gptqaccurateposttrainingquantization, liu2024qllm} and run-time activation sparsity \citep{teal, wang-etal-2025-weight, sparsing-law, hou2026acttailglobalactivationsparsity} reduce memory and latency costs, they leave the underlying structural redundancy intact. Model pruning addresses this directly. Unstructured methods achieve excellent performance retention at high sparsity ratios \citep{wanda, sparsegpt}, but sparsity alone does not yield practical speed-ups without hardware-specific sparse kernels. This has motivated a shift toward structured pruning, which removes contiguous parameter groups such as neurons \citep{ma2023llmpruner, ashkboos2024slicegptcompresslargelanguage, prunenet}, MLP blocks \citep{zhong-etal-2025-blockpruner, zhang2024finercutfinergrainedinterpretablelayer}, attention heads \citep{michel2019sixteenheadsreallybetter, chen-etal-2025-alps, fairness-aware-structured-pruning}, or entire layers \citep{men-etal-2025-shortgpt, song2024sleb, shopkhoev2026replacemenetworksimplificationdepth}, yielding models that are smaller, leaner, and faster without specialized kernel support.

Despite these advances, progress has been largely confined to dense Transformer architectures. Many existing methods either lack support for MoE models entirely or naively transpose dense pruning heuristics onto MoE architectures. For instance, MoE-Pruner \citep{xie2024moeprunerpruningmixtureofexpertslarge} extends Wanda's weight-magnitude-times-activation heuristic \citep{wanda} to MoE models by simply multiplying it with router weights. While this incorporates a MoE-specific signal, it overlooks the deeper structural properties and inter-component dependencies that arise from an expert-centric design. This has motivated the emergence of MoE-centric pruning methods \citep{lu-etal-2024-expert-sparsity, lee-etal-2025-stun, mosaic-pruning, lasby2026reap} that propose heuristics such as activation variability scores \citep{mosaic-pruning} and router-weighted expert norms \citep{lasby2026reap} to better capture architectural nuances. While effective, these methods suffer from a critical drawback: they use locally derived heuristics to drive global pruning decisions. We hypothesize that such myopic heuristics are bound to overlook important global dependencies, leading to concessions in post-compression performance.

Beyond pruning, expert merging has emerged as a complementary strategy for compressing MoE models. Rather than discarding experts outright, merging methods \citep{he-etal-2023-merging-meo, chen2025hcsmoe, li2024mergecompressdemystifyefficient, muqeeth2024softmergingexpertsadaptive} combine redundant experts into a smaller set, preserving the collective knowledge encoded across them. For instance, MEO \citep{he-etal-2023-merging-meo} employs router-weighted averaging of expert weights to merge experts based on routing similarity. In contrast, HC-SMoE \citep{chen2025hcsmoe} leverages hierarchical clustering to merge functionally similar experts. While effective, \citet{lasby2026reap} have recently shown that expert merging methods introduce an \textit{irreducible error} due to loss of fine-grained router control, serving as a strong motivator for exploring increasingly advanced pruning strategies. 

% To tackle the locality limitation of prior pruning work, we directly incorporate inter-expert dependencies within our saliency heuristic, leading to globally-aware pruning configurations.

\section{Methodology}
\label{sec:methodology}
\begin{figure*}[t]
    \centering
    {\includegraphics[width=0.95\linewidth]{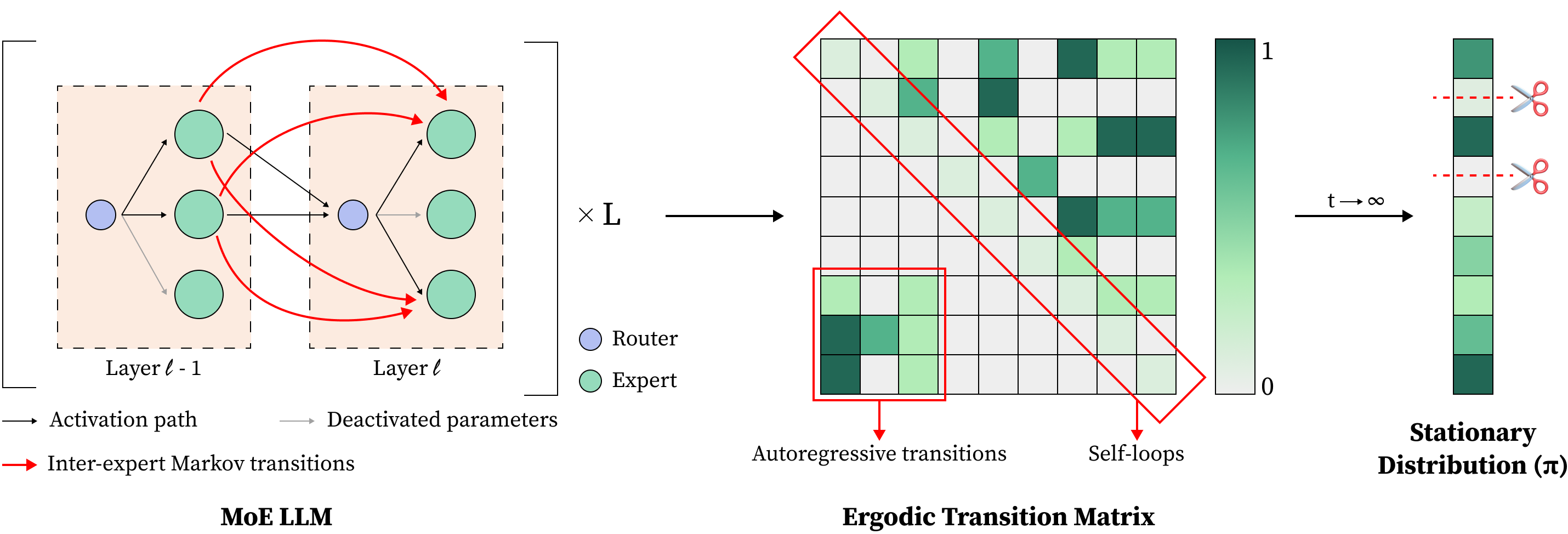}}
    \caption{Schematic of \pruner's methodology. \pruner\ visualizes expert routing in MoE LLMs in the form of an Ergodic Markov chain and uses its stationary distribution as a global heuristic to remove redundant experts from each layer.}
    \label{fig:schematic}
\end{figure*}

Figure~\ref{fig:schematic} displays our proposed calibration-driven procedure for structurally pruning MoE language models.  
The central idea is to formulate top-$k$ expert routing within the model as a Markov chain over all states of the form $(\texttt{layer}, \texttt{expert})$ and to identify and remove redundant experts as those whose steady-state probabilities under this chain are smallest. Since expert tensors are physically sliced rather than masked at inference-time, the resulting model is strictly smaller in overall parameter count and memory footprint. 
% In accordance with existing literature, we employ a LoRA-based recovery fine-tuning (RFT) stage that restores task performance lost through pruning while preserving the chain-derived structural decisions.

\subsection{Preliminaries and Problem Setup}
\label{sec:method:setup}
Let $\mathcal{M}$ denote a MoE model with $L$ layers, each containing a set of $E$ experts $\{f_{\ell,1}, \dots, f_{\ell,E}\}$ and a router $g_\ell : \mathbb{R}^{d} \to \mathbb{R}^{E}$. For an input token representation $h \in \mathbb{R}^{d}$ at layer $\ell$, the router produces routing weights $r_\ell$ where the routing weight for expert $e$ is computed as follows:
\begin{align}
\label{eq:routing-weights}
    r_\ell^{(e)} = \frac{\exp\big[g_\ell(h)_e\big]}{\sum_{e'}\exp\big[g_\ell(h)_{e'}\big]}, \quad 1\leq e'\leq E
\end{align}
The router's top-$k$ operator selects an index set $\mathcal{E}_\ell(h) \subset \{1,\dots,E\}$ with $|\mathcal{E}_\ell(h)| = k$. The layer output, $A_\ell$, is then the weighted sum over the selected experts, using the routing weights from Equation~\ref{eq:routing-weights} re-normalized over $\mathcal{E}_\ell(h)$. Formally,
\begin{align}
    A_\ell = &\sum_{e \in \mathcal{E}_\ell(h)} \hat{r}_\ell^{(e)}(h)\, f_{\ell,e}(h) 
    \\\text{where} \quad &\hat{r}_\ell^{(e)}(h) = \frac{r_\ell^{(e)}}{\sum_{e' \in \mathcal{E}_\ell(h)} r_\ell^{(e')}}
\end{align}
Given a target compression ratio $\rho \in (0,1)$, our objective is to select a set $\mathcal{R} \subset \{(\ell, e) : 1 \le \ell \le L,\, 1 \le e \le E\}$ of $K = \lfloor \rho \cdot L E \rfloor$ expert slots to delete from $\mathcal{M}$, yielding a pruned model $\tilde{\mathcal{M}}$.

\subsection{Routing as a Markov Chain over Experts}
\label{sec:method:markov}

Let $\mathcal{S} = \{(\ell, e) : 1 \le \ell \le L,\, 1 \le e \le E\}$ index the $LE$ expert slots in the model. As a token traverses the network, it activates a sequence of states in $\mathcal{S}$. Inspired by Markovian principles, we propose that, conditioned on the experts selected at layer $\ell$, the experts selected at layer $\ell{+}1$ are governed by a transition kernel that depends only on the layer index and the current selection. This memoryless assumption, motivated by the fact that the router at layer $\ell{+}1$ consumes only the residual stream produced by layer $\ell$, induces a Markov chain over $\mathcal{S}$. In order to derive a single steady-state (stationary) probability distribution of this chain, we induce Ergodicity in it, \textit{i.e.}, ensure irreducibility (all states are reachable) and aperiodicity (not all paths between any two states have the same period). While the former is achieved 
% To obtain a single chain whose steady-state probability distribution can be analyzed, we induce we close the chain cyclically 
by connecting the final layer back to the first layer, the latter is the resultant of a smoothing technique that applies a small self-loop to each state. The resulting chain admits a unique stationary distribution $\pi$, which can be interpreted as the long-run probability with which routing visits each expert slot. Experts whose stationary mass is smallest contribute least to overall computation and are therefore the natural targets for removal. Since $\pi$ is derived from long token trajectories across the model's layers, it encodes long range inter-expert dependencies and structural routing patterns which are often invisible to local heuristics. 

\subsection{Calibration and Transition-Count Estimation}
\label{sec:method:calibration}

We estimate the per-layer transition kernels empirically from a small calibration corpus. 
% By default we use $N$ SlimOrca prompts at context length $512$, drawn deterministically from the head of the dataset; alternative corpora (Wikitext, LAMBADA, C4, Alpaca) are supported and ablated.
For each calibration batch, the model is run \textit{autoregressively} for $T_{seq}$ generation steps, with each predicted token appended to the rolling context. Notably, in contrast to teacher-forced passes, autoregressive rollouts ensure that transition statistics reflect the distribution the model itself induces over expert activations during inference.

% For each layer $\ell$, we empirically count all transitions of the form $f_{\ell-1, i} \rightarrow f_{\ell, j}$, representing the activation of expert $j$ of layer $\ell$ after expert $i$ from layer $\ell-1$ was activated by the same token. 
Let $\mathbf{e}_\ell \in \mathbb{Z}^{B \times T_\text{seq} \times k}$ represent the top-$k$ index tensor produced by the router at layer $\ell$. We accumulate pairwise transition counts from experts in layers $\ell-1$ to those in layer $\ell$ into a per-layer matrix $C^{(\ell)} \in \mathbb{R}^{E \times E}$. 
% To collect transition statistics without modifying the model definition, we instrument every MoE layer with two lightweight forward hooks. The router hook captures the top-$k$ index tensor produced at layer $\ell$ into a per-step buffer $\mathbf{e}_\ell \in \mathbb{Z}^{B \times T_\text{seq} \times k}$. Immediately afterwards, the MLP hook, executed after the original forward pass so as not to perturb the model's outputs, reads both $\mathbf{e}_{\ell-1}$ (the previous-layer selection cached by the prior hook firing) and $\mathbf{e}_{\ell}$, and accumulates pairwise transition counts into a per-layer matrix $C^{(\ell)} \in \mathbb{R}^{E \times E}$. 
Specifically, $C^{(\ell)}_{ij}$ counts the number of token-steps for which expert $i$ 
was selected at layer $\ell{-}1$ and expert $j$ was selected at layer $\ell$. Formally,

\begin{equation}
    C^{(\ell)}_{ij} = \sum_{b=1}^{B} \sum_{t=1}^{T_{\text{seq}}} 
    \mathbf{1}\!\left[i \in \mathbf{e}_{\ell-1}^{(b,t)}\right] \cdot 
    \mathbf{1}\!\left[j \in \mathbf{e}_{\ell}^{(b,t)}\right],
\end{equation}
where the sum ranges over all \textit{non-padding} tokens $(b, t)$ across the batch.
% Concretely, for every token position we enumerate all $k^2$ source--target index pairs $(e_i, e_j)$ with $e_i \in \mathbf{e}_{\ell-1}$ and $e_j \in \mathbf{e}_\ell$, linearise them as $e_i \cdot E + e_j$, and accumulate via \texttt{torch.bincount}, vectorising the count update across the entire batch and sequence in a single kernel call. 
These per-layer matrices are accumulated across all batches and autoregressive steps and cached to disk so subsequent runs can re-use the calibration statistics.

\subsection{Global Transition Matrix and Stationary Distribution}
\label{sec:method:steady}

We combine the per-layer count matrices into a single global matrix $\mathbf{T}$,
which we partition into an $L \times L$ grid of $E \times E$ blocks. 
% Only $L$ of these blocks are non-zero since they connect each layer to the next. 
Mathematically,
\begin{equation}
    \mathbf{T}[\ell,\, \ell'] =
    \begin{cases}
        C^{(\ell')} & \text{if } \ell' = \ell + 1 \bmod L \\
        0                    & \text{otherwise}
    \end{cases}
\end{equation}
where $\mathbf{T}[\ell, \ell']$ represents the block at block-index $(\ell, \ell')$. More explicitly, 
\begin{equation}
    \mathbf{T} =
    \begin{pmatrix}
        0          & C^{(1)}    &        &             \\
                   & 0          & \ddots &             \\
                   &            & \ddots & C^{(L-1)}   \\
        C^{(0)}    &            &        & 0
    \end{pmatrix}
\end{equation}
The final block $\mathbf{T}[L{-}1,\, 0] = C^{(0)}$
closes the cycle. The autoregressive nature of the model, wherein the token generated at the last layer is fed to the first layer at the next timestep, makes this cyclic closure a natural choice, allowing us to capture inter-expert transitions based on the model's inherent distribution and behavior. 
% , allowing us to treat the layer sequence as periodic 
% Therefore, rather than having a chain that terminates at the last layer, we leverage its autoregressive properties and treat the layer sequence as periodic, 
% so that standard Markov chain analysis applies without any boundary conditions.

We then row-normalize $\mathbf{T}$ to obtain a stochastic matrix $\mathbf{P}$:
\begin{equation}
\mathbf{P}_{ij} \;=\; \frac{\mathbf{T}_{ij}}{\max\!\bigl(1,\,\sum_{j'} \mathbf{T}_{ij'}\bigr)},
\end{equation}
where the denominator is clipped to one to handle expert slots that are never activated on the calibration set. The empirical chain, while irreducible due to its autoregressive modeling, is periodic. To tackle this, we apply an $\varepsilon$-smoothing operator that adds a small self-loop to every state,
\begin{equation}
\label{eq:aperiodic-transition-matrix}
\mathbf{P}_{\!\varepsilon} \;=\; \frac{(1-\varepsilon)\,\mathbf{P} + \varepsilon\,\mathbf{I}}{Z},
\end{equation}
where $\varepsilon$ is the hyperparameter that controls this smoothing and $Z$ is the row-sum re-normalizer. The resulting kernel is irreducible and aperiodic on the support of the calibration trajectories, so the Perron--Frobenius theorem guarantees a unique stationary distribution $\pi$ satisfying $\pi^\top \mathbf{P}_{\!\varepsilon} = \pi^\top$. We compute $\pi$ by power iteration as follows: 
\begin{equation}
\pi_{t+1}^{\top} \;=\; \pi_t^{\top}\, \mathbf{P}_{\!\varepsilon},
\end{equation}
where $\pi_0 = \tfrac{1}{LE}\mathbf{1}$ (uniform initial probabilities) and the iteration terminates when $\|\pi_{t+1} - \pi_t\|_1 < \tau$ or after $N_\tau$ iterations, whichever condition is hit first.

\subsection{Expert Selection}
\label{sec:method:selection}

% Given the stationary distribution $\pi$, we rank the $LE$ expert slots in ascending order of $\pi$-mass and mark the bottom $K = \lfloor \rho \cdot L E \rfloor$ slots as redundant.
Let $\sigma_\ell$ be the permutation that sorts all expert slots of layer $\ell$ in ascending order 
of $\pi$-mass. We mark the bottom $K_\ell = \lfloor \rho \cdot E \rfloor$ experts as redundant:
\begin{equation}
    \mathcal{R_\ell} = \{\sigma_\ell(1),\, \sigma_\ell(2),\, \ldots,\, \sigma_\ell(K_\ell)\}.
\end{equation}
The selected set of redundant experts, $\mathcal{R_\ell}$, is chosen as a consequence of the model's global routing patterns under a uniformity constraint that ensures no layer is pruned overly aggressively.   
% across all layers; the number of experts removed from any individual layer is a consequence of the data and is not constrained \emph{a priori}, allowing the procedure to adapt the per-layer pruning rate to each layer's routing diversity. 
% As a control, we also implement a random baseline that draws $K$ pairs uniformly without replacement from $\mathcal{S}$; we use this baseline to isolate the contribution of the Markov-chain importance signal from the structural pruning machinery itself.

\subsection{Structural Removal of Redundant Experts}
\label{sec:method:removal}

% Let $\mathcal{R}_\ell = \{e : (\ell, e) \in \mathcal{R}\}$ denote the pruned slots in layer $\ell$ whose corresponding parameters are to be removed. 
Let $\bar{\mathcal{E}}_\ell = \{1,\dots,E\} \setminus \mathcal{R}_\ell$ denote the retained experts in layer $\ell$. We slice the expert-bank tensors along the expert dimension as follows:
\begin{align}
\mathbf{W}^{(i)}_{\ell} &\leftarrow \mathbf{W}^{(i)}_{\ell}[\bar{\mathcal{E}}_\ell,:,:] \\
% \mathbf{W}^{\text{down}}_{\ell} &\leftarrow \mathbf{W}^{\text{down}}_{\ell}[\bar{\mathcal{E}}_\ell,:,:]
\mathbf{b}^{(i)}_{\ell} &\leftarrow \mathbf{b}^{(i)}_{\ell}[\bar{\mathcal{E}}_\ell,:]
\end{align}
where $\mathbf{W}^{(i)}_{\ell}$ is the $i^{th}$ expert-bank tensor in layer $\ell$ and $\mathbf{b}^{(i)}_{\ell}$ is its corresponding bias. The router weight $\mathbf{W}^{\text{router}}_\ell \in \mathbb{R}^{E \times d}$ has its rows restricted to $\bar{\mathcal{E}}_\ell$, eliminating the logits associated with deleted experts. 
% Sliced tensors are re-wrapped as \texttt{nn.Parameter} objects and re-attached to the layer module; 
% The per-layer attributes \texttt{num\_experts} (or \texttt{n\_routed\_experts}, depending on architecture) are updated accordingly, and the
The router's top-$k$ parameter is clamped to $\min(k, |\bar{\mathcal{E}}_\ell|)$ so that selection remains well-defined in layers where the survivor count has fallen below the original $k$. Owing to the structural removal of experts, the pruned model contains strictly fewer parameters and consumes proportionally less memory without requiring additional masking or run-time bookkeeping to realize these gains. 
% After pruning, the temporary forward hooks installed for calibration are removed and the original, kernel-optimised expert and router forwards are restored, ensuring that inference uses the fused MoE kernels of the underlying framework.

\subsection{Recovery Fine-Tuning}
\label{sec:method:rft}
Aggressive pruning unavoidably perturbs the model's output distribution. We recover task performance with a brief LoRA-based recovery fine-tuning stage \citep{hu2021loralowrankadaptationlarge}. 
% LoRA adapters of rank $r{=}16$ and scaling $\alpha{=}16$ (no dropout, no bias adaptation) are inserted on the attention projections $\{\mathbf{W}^{Q}, \mathbf{W}^{K}, \mathbf{W}^{V}, \mathbf{W}^{O}\}$ of every transformer block. 
Crucially, LoRA adapters are inserted on each layer's attention projections ($\{\mathbf{W}^{Q}, \mathbf{W}^{K}, \mathbf{W}^{V}, \mathbf{W}^{O}\}$) while the expert weights and router parameters are kept frozen. This ensures that the structural decisions made by the Markov-chain analysis are preserved; the adapter learns to compensate by reshaping attention behavior around the surviving experts rather than by re-mixing the routing distribution.

% \subsection{Architecture-Agnostic Implementation}
% \label{sec:method:generality}

% The above procedure is defined entirely in terms of the abstract MoE primitives---a router producing top-$k$ indices and an expert bank applying expert-indexed FFNs---and is therefore agnostic to specific model families. We apply the identical algorithm to GPT-OSS, Qwen3-MoE, and Nemotron-3 \todo{Decide whether nemotron is a part of the final list or not} by abstracting away architecture-specific details such as attribute names and expert-bank layouts. 
% % (i) the attribute path used to reach the router (e.g.\ \texttt{layer.mlp.router} in GPT-OSS and Llama-4, \texttt{layer.mlp.gate} in Qwen and DeepSeek, \texttt{layer.mixer.gate} in Nemotron-H), (ii) the layout of the expert bank (a dense $E$-stacked tensor in GPT-OSS, Qwen3.5 and Nemotron-H, an \texttt{nn.ModuleList} of per-expert sub-modules in Qwen3), and (iii) the subset of decoder layers that are MoE-bearing (Nemotron-H interleaves MoE and dense blocks, identified via \texttt{config.layers\_block\_type}). 
% The Markov-chain construction, stationary-distribution computation, ranking, and structural slicing are unchanged across all families, demonstrating that the proposed criterion is a general property of MoE routing rather than an artifact of any single architecture.

% ---------------------- ORIGINAL -----------------------------------
\begin{table*}[!htb]
\centering
\resizebox{\textwidth}{!}{%
\begin{tabular}{l||clccccc||cc}
\hline
\textbf{Model} & \textbf{CR} & \textbf{Method} & \textbf{Generative} & \textbf{World} & \textbf{Domain} & \textbf{NLU \& NLI} & \textbf{Safety} & \textbf{Avg RP (\%)} & \textbf{Std RP (\%)} \\
 & & & (Log-PPL $\downarrow$) & (Acc $\uparrow$) & (Acc $\uparrow$) & (Acc $\uparrow$) & (Score $\uparrow$) & ($\uparrow$) & ($\downarrow$) \\
\hline
%-----------------------------------------------
% GPT-OSS
\multirow{9}{*}{GPT-OSS-20B} & \multirow{1}{*}{-} & Base & 2.17 & 65.62 & 56.82 & 74.16 & 58.17 & - & - \\
\cdashline{2-10}
& \multirow{5}{*}{25\%} & Random & 2.39 &  63.60 & 52.53 & 72.40 & 58.66 & 95.38 & 6.00\\
 % & Frequency & 2.21 & 64.35 & 56.05 & 73.71 & 57.18 & 97.76 & 2.10\\
 & & HC-SMoE & 2.54 & 57.86 & 47.07 & 72.55 & 57.26 & 90.46 & 8.60\\
 & & MOP & 2.31 & 63.99 & 53.04 & 72.85 & 56.59 & 95.63 & 5.10\\
 & & REAP & 2.21 & 64.19 & 55.46 & 73.57 & 57.52 & 97.67 & 2.70\\
 & & \pruner & 2.18 & 64.68 & 57.30 & 73.92 &  57.97 & \textbf{98.90} & \textbf{1.90}\\
\cdashline{2-10}
% 50% COMPRESSION
& \multirow{5}{*}{50\%} & Random & 2.95 & 52.35 & 44.78 & 65.21 & 57.82 & 83.92 & 11.50\\
% & Frequency & 2.45 & 59.77 & 50.43 & 71.29 & 55.67 & 91.71 & 6.70\\
& & HC-SMoE & 3.00 & 56.39 & 39.26 & 69.17 & 56.24 & 83.71 & 12.50\\
& & MOP & 2.56 & 60.06 & 43.06 & 71.90 & 54.85 & 88.41 & 10.80\\
& & REAP & 2.47 & 58.49 & 44.70 & 71.17 & 56.62 & 89.37 & 10.40\\
& & \pruner & 2.40 & 60.64 & 50.61 & 71.84 & 56.14 & \textbf{92.59} & \textbf{6.30}\\
\cline{1-10}
%-----------------------------------------------
% QWEN
\multirow{9}{*}{Qwen-3-30B} & \multirow{1}{*}{-} & Base & 1.97 & 69.43 & 63.29 & 76.65 & 56.60 & - & -\\
\cdashline{2-10}
% 25% COMPRESSION
& \multirow{4}{*}{25\%} & Random & 2.19 & 62.67 & 53.57 & 74.49 & 54.00 & 91.79 & 7.10\\
 % &  & Frequency & 1.87 & 70.56 & 61.93 & 76.44 & 53.67 & 99.63 & 3.90\\
 &  & HC-SMoE & 2.35 & 59.62 & 46.65 & 72.06 & 56.20 & 87.47 & 11.00\\
 % &  & MOP & -\\
 &  & REAP & 1.93 & 68.86 & 59.77 & 76.58 & 54.26 & 98.13 & 4.30\\
 &  & \pruner & 1.86 & 70.00 & 61.56 & 76.47 & 53.95 & \textbf{99.40} & \textbf{3.50}\\
\cdashline{2-10}
% 50% COMPRESSION
& \multirow{4}{*}{50\%} & Random & 2.86 & 54.65 & 44.97 & 61.06 & 54.18 & 78.77 & 13.50\\
% &  & Frequency & 2.45 & 59.77 & 50.43 & 71.29 & 55.67 & 91.71 & 6.70\\
&  & HC-SMoE & 3.23 & 47.37 & 36.61 & 65.54 & 53.70 & 74.58 & 16.60\\
% &  & MOP & -\\
&  & REAP & 2.45 & 63.56 & 53.54 & 72.48 & 54.45 & 90.25 & \textbf{7.10}\\
&  & \pruner & 2.04 & 65.82 & 52.07 & 75.71 & 53.81 & \textbf{93.40} & 10.50 \\
\cline{1-10}
%------------------------------------------------
\end{tabular}%
}
\caption{Post-compression performance of GPT-OSS-20B and Qwen3-30B at compression ratios of 25\% and 50\% after undergoing RFT. \pruner\ demonstrates excellent performance consistently across diverse MoE-based architectures. Best results are highlighted in boldface.}
\label{tab:gpt-qwen-main-results}
\end{table*}

\section{Experimental Setup}

\paragraph{Model Families.} 
We benchmark \pruner\ on two different MoE-based models, namely, GPT-OSS-20B\footnote{\url{https://huggingface.co/openai/gpt-oss-20b}} \citep{openai2025gptoss120bgptoss20bmodel} and Qwen-3-30B\footnote{\url{https://huggingface.co/Qwen/Qwen3-30B-A3B}}  \citep{yang2025qwen3technicalreport}. Specifically, GPT-OSS-20B model contains 32 experts per layer in contrast to 128 experts in Qwen-3-30B, allowing us to evaluate \pruner\ under both, expert-sparse and expert-dense routing paradigms.
% Additionally, Nemotron consists of Mamba-2 layers \citep{mamba2} interleaved between MoE layers, making it a hybrid architecture that tests \pruner's ability to generalize beyond pure Transformer-based MoE designs.

\paragraph{Baselines.} 
We compare \pruner\ against a naive, random pruning baseline and three contemporary baselines: an expert merging framework, HC-SMoE \citep{chen2025hcsmoe}, and two MoE-pruning methods, Mosaic Pruning (MoP) \citep{mosaic-pruning} and REAP \citep{lasby2026reap}. All baselines were manually adapted to support the evaluated model families and our evaluation framework. To ensure a fair comparison, all configurations are held constant across baselines, with only the core pruning methodology varying. 
% We provide all implementation details in Appendix~\ref{appendix:implementation-details}. 
Note that MoP holds each expert's output for every calibration token in memory. While this is manageable for GPT-OSS-20B (32 experts per layer), it requires a prohibitively high amount of memory for Qwen-3-30B (128 experts per layer). As a result, we are unable to run MoP on Qwen and report its results only for GPT OSS. 
% We evaluate \pruner\ against state-of-the-art baselines: MoP \citep{mosaic-pruning} and Expert Sparsity \citep{lu-etal-2024-expert-sparsity}. 
% MoP aims to identify a set of functionally holistic experts via a ``cluser-then-select'' methodology while Expert Sparsity employs model performance deviation as a direct proxy for expert utility. Both baselines have been extended to support the model families used in our evaluation.

\paragraph{Evaluation Tasks.} We evaluate all methods on a comprehensive testbed of 17 tasks across five domains. We measure generative aptitude by computing log-perplexity on the WikiText-2 \citep{wikitext} and LAMBADA \citep{paperno2016lambada} datasets. PIQA \citep{bisk2020piqa}, PROST \citep{aroca-ouellette-etal-2021-prost}, and CommonsenseQA \citep{talmor-etal-2019-commonsenseqa} are employed to gauge world understanding and common sense reasoning. We assess domain-specific capabilities via science-based reasoning tasks such as ARC-Easy and ARC-Challenge \citep{clark2018thinksolvedquestionanswering}, and a medical question-answering dataset called MedQA \citep{medqa}. Additionally, we employ OpenbookQA \citep{OpenBookQA2018} for multi-domain advanced question-answering and reasoning. We use the BLiMP \citep{warstadt2020blimp}, BoolQ \citep{clark2019boolq}, LAMBADA \citep{paperno2016lambada}, Winogrande \citep{sakaguchi2021winogrande}, and COQA \citep{reddy2019coqa} benchmarks to quantify natural language understanding and inference capabilities. Notably, unlike prior MoE baselines, we assess models from a lens of Safety, Bias, and Ethics via popular benchmarks such as Winogender \citep{rudinger-etal-2018-winogender}, TruthfulQA \citep{lin-etal-2022-truthfulqa}, and Moral Stories \citep{emelin-etal-2021-moral-stories}.

\paragraph{Implementation Details.} \pruner\ has been implemented in PyTorch \citep{paszke2019pytorchimperativestylehighperformance} using Huggingface's Transformers library \citep{wolf-etal-2020-transformers}. We conduct all experiments on a cloud-based GPU server consisting of 8 NVIDIA A100 GPUs, each comprising of 80GB of VRAM. For post-compression recovery fine-tuning (RFT), we employ the PEFT library \citep{peft} and conduct evaluation using the Language Model Evaluation Harness \citep{lm-eval-harness}. 

We estimate expert transition counts (Section~\ref{sec:method:calibration}) by calibrating the model on 250 samples of the SlimOrca dataset \citep{SlimOrca}, each with a context length of 512 tokens, achieved via padding or truncation. All transitions are counted over $T_{seq} = 100$ autoregressive generation steps. 
Notably, we utilize masking to ensure that padding tokens are not considered when counting transitions. In order to induce aperiodicity in the resulting stochastic transition matrix, we employ $\varepsilon$-smoothing, where $\varepsilon = 10^{-5}$ (Equation~\ref{eq:aperiodic-transition-matrix}). In order to compute the stationary distribution of our final transition matrix, we employ power iteration with a tolerance of $\tau = 10 ^ {-6}$ under a maximum of $N_\tau = 1000$ (Section~\ref{sec:method:steady}) iterations to balance accuracy with computational efficiency.

After compression, we conduct RFT on 2{,}000 SlimOrca conversations with 1{,}024 tokens each, using LoRA adapters of rank $r{=}16$ and scaling $\alpha{=}16$ (no dropout) that are inserted on the attention projections of each layer.
Training is done over a single epoch using the AdamW optimizer \citep{adamw} with a peak learning rate of $2\times 10^{-4}$, a linear warm-up over the first $10\%$ of steps followed by linear decay, and a batch size of $2$.

\begin{figure*}[t]
    \centering
    {\includegraphics[width=0.9\linewidth]{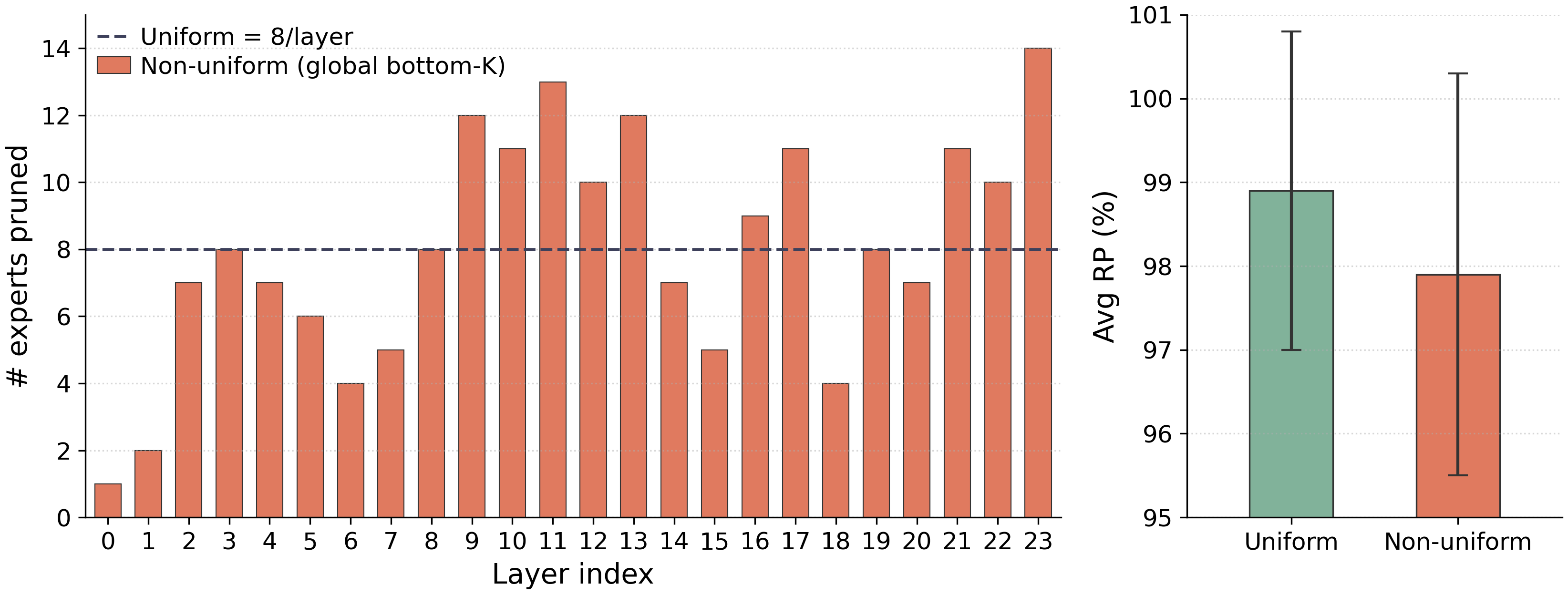}}
    \caption{Comparison of uniform vs.\ non-uniform expert pruning strategies applied to GPT-OSS-20B at 25\% compression. Uniform pruning (8 experts/layer) achieves a higher average retained performance (RP) compared to non-uniform pruning, indicating that allowing unconstrained global pruning can over-prune sensitive layers and hurt overall model quality, despite the flexibility it affords.}
    \label{fig:uniform-non-uniform-comparison}
\end{figure*}
% ---------------------------------------------------------
\section{Results}
Table~\ref{tab:gpt-qwen-main-results} provides quantitative performance metrics 
of various pruning methods on two architecturally 
distinct MoE models: GPT-OSS-20B and Qwen3-30B. We find \pruner\ to be consistently superior across both architectures, demonstrating that it is not a method tuned to a single model family but rather a general-purpose pruner whose Markov-chain formulation transfers across diverse MoE topologies. On GPT-OSS-20B, \pruner\ outperforms the best baseline (REAP) by relatively $1.26\%$ at 25\% compression and by $3.60\%$ at 50\% compression. On Qwen3-30B, the relative margins are equally strong: $1.29\%$ over REAP at 25\% and $3.49\%$ at 50\% compression. Even under a strict 50\% compression regime, \pruner\ surpasses HC-SMoE by up to $10.61\%$ on GPT-OSS-20B and by $25.23\%$ on Qwen3-30B, underscoring its robustness under aggressive compression.

\paragraph{Task-level analysis.}
A closer inspection of individual tasks reveals that the gains of \pruner\ are 
not uniform but are most pronounced in the Generative and Domain dimensions, 
which are arguably the most sensitive indicators of expert specialization. On GPT-OSS-20B at 25\% compression, \pruner\ 
achieves a Domain accuracy of $57.30$, which not only surpasses every baseline 
but also \emph{exceeds the unpruned base model} ($56.82$), suggesting that 
removing low-salience experts can act as a mild regularizer by eliminating 
redundant or noisy routing paths. Its Log-PPL of $2.18$ is within $0.01$ of the 
base ($2.17$), effectively recovering near-lossless generative quality that no 
baseline achieves. A strikingly similar pattern emerges on Qwen3-30B: at 25\% 
compression, \pruner\ achieves a Log-PPL of $1.86$, which \emph{beats} the 
unpruned base model ($1.97$), and a World Knowledge accuracy of $70.00$ that 
again \emph{surpasses} the base ($69.43$). These results, unlikely to be coincidental, corroborate our central claim that modeling expert routing as an Ergodic Markov chain and 
retaining experts with the highest stationary probabilities per layer preserves the experts that collectively carry the most global information flow, rather than those that merely appear frequently in local or layer-wise snapshots while ensuring no layer is over-pruned. We provide an analysis of per-task model performance in Appendix~\ref{app:detailed-results}.

\paragraph{Graceful degradation under aggressive compression.}
A notable property of \pruner\ is its graceful degradation as the compression 
ratio increases. On 
GPT-OSS-20B, moving from 25\% to 50\% compression, \pruner\ incurs an average RP drop of only $6.31$ percentage points, compared to drops of $7.22$ (MoP), $6.75$ (HC-SMoE), $11.46$ (Random), and $8.30$ (REAP). This trend can be observed on Qwen3-30B as well: \pruner\ drops by just $6.00$ percentage points, versus $13.02$ (Random), $12.89$ (HC-SMoE), and $7.88$ (REAP). This suggests that the stationary distribution provides a well-calibrated, architecture-agnostic global ranking of 
expert importance: even when a large fraction of experts is removed, the retained set still faithfully represents the model's aggregate routing behavior regardless of the underlying MoE topology. Notably, on GPT-OSS-20B, \pruner\ at 50\% compression ($92.59\%$ RP) still outperforms HC-SMoE at the much lighter 25\% setting ($90.46\%$ RP); the analogous gap is even wider on Qwen3-30B, where \pruner\ at 50\% ($93.40\%$) eclipses HC-SMoE at 25\% ($87.47\%$) by $6.78\%$. 

\paragraph{Consistency as a proxy for routing coherence.}
Beyond raw performance, the standard deviation of RP serves as a proxy for how well a pruned model generalizes across heterogeneous tasks. On GPT-OSS-20B, 
\pruner\ achieves the lowest Std RP at both compression ratios ($1.90\%$ at 25\% 
and $6.30\%$ at 50\%), compared to the next best competitor at each level (REAP: 
$2.70\%$; REAP: $10.40\%$, respectively). On Qwen3-30B at 25\% compression, 
\pruner\ again records the lowest standard deviation ($3.50\%$) versus REAP 
($4.30\%$), Random ($7.10\%$), and HC-SMoE ($11.00\%$). At 50\% compression on 
Qwen3-30B, REAP achieves a marginally lower Std RP ($7.10\%$) compared to 
\pruner\ ($10.50\%$), though \pruner\ retains a substantially higher average RP 
($93.40\%$ vs. $90.25\%$), indicating that the small consistency advantage of 
REAP comes at the cost of significantly greater performance loss. The 
improvement in consistency is particularly striking at 50\% compression on 
GPT-OSS-20B, where competing methods exhibit standard deviations of $10$--$12\%$, roughly twice that of \pruner. Local heuristics such as the ones used by baseline methods tend to over-rely on task-dominant experts, often turning a blind eye to experts that are moderately active across various tasks and anchor cross-task consistency. This structural blind spot manifests in the form of high variance. By leveraging a globally relevant importance score, \pruner\ bypasses this issue, yielding models that are much more consistent across diverse benchmarks. 

\begin{table*}[t]
\centering
% Ablations
\resizebox{\textwidth}{!}{%
\begin{tabular}{cccccc|cc}
\hline
\textbf{Variant} & \textbf{Generative} & \textbf{World} & \textbf{Domain} & \textbf{NLU \& NLI} & \textbf{Safety} & \textbf{Avg RP (\%)} & \textbf{Std RP (\%)} \\
& (Log-PPL $\downarrow$) & (Acc $\uparrow$) & (Acc $\uparrow$) & (Acc $\uparrow$) & (Score $\uparrow$) & ($\uparrow$) & ($\downarrow$) \\
\hline
% 50 samples
\pruner & 2.18 & 64.68 & 57.30 & 73.92 &  57.97 & \textbf{98.90} & \textbf{1.90}\\
\cdashline{1-8}
\multicolumn{1}{r}{\textsc{-- uni}} & 2.19 & 63.85 & 55.78 & 74.03 & 57.37 & 97.90 & 2.40\\
\multicolumn{1}{r}{\textsc{-- auto}} & 2.21 & 64.35 & 56.05 & 73.71 & 57.18 & 97.76 & 2.10\\
% \textsc{--PAD\_MASK} & 2.19 & 64.71 & 56.65 & 74.04 & 57.82 & 98.62 & 1.80\\
\multicolumn{1}{r}{\textsc{-- rft}} & 5.12 & 59.42 & 55.56 & 47.85 & 57.28 & 78.54 & 26.40\\
\textsc{REAP -- rft} & 5.02 & 59.60 & 52.75 & 48.67 & 57.06 & 77.91 & 25.60\\
\hline
\end{tabular}%
}
\caption{Ablation results for \pruner\ obtained after compressing GPT-OSS-20B by 25\%.}
\label{table:ablations}
\end{table*}
\paragraph{Ablation Study.} Table~\ref{table:ablations} provides results of various ablative variants of \pruner, allowing for a deeper analysis of the necessity and effectiveness of its components. We analyze these variants as follows. 

\begingroup
\leftskip=1em
\noindent\textsc{\textbf{-- uni.}} This variant removes the uniform pruning constraint from \pruner\ and prunes experts with the globally minimum stationary probabilities. As analyzed in Section~\ref{sec:discussion}, while removing the uniformity constraint leads to a variant that is more flexible and can dynamically allocate the pruning budget across layers, it can also lead to configurations where certain layers are over-pruned, a bottleneck that reflects clearly in the model's post-compression performance metrics ($97.90\%$ v/s $98.90\%$ avg RP) and task-wise consistency ($26.32\%$ higher standard deviation than base variant). Empirically, ensuring the stability of the pruning process leads to better overall performance even if it comes at the cost of decisional flexibility, motivating us to the adopt the uniform regime. 

\endgroup

\begingroup
\leftskip=1em
\noindent\textsc{\textbf{-- auto.}} This variant ablates the autoregressive expert transition counting component from \pruner. Consequently, \pruner's Markov modeling is performed using single-pass activation heuristics which fail to capture the model's long range routing behavior (Section~\ref{sec:discussion}). As a result, this variant incurs a relative drop of $1.15\%$ in average performance, along with a $10.53\%$ increase in standard deviation across tasks, demonstrating the effectiveness of autoregressive routing trajectories at aiding cross-task generalization and consistency.

\endgroup

\begingroup
\leftskip=1em
\noindent\textsc{\textbf{-- rft.}} We also explore the effectiveness of performing post-compression RFT via the \textsc{-- rft} variant which ablates this exact process. We observe that this variant performs considerably worse than \pruner's base variant on average performance retention ($20.59\%$ relative drop) as well as cross-task consistency ($1,289.47\%$ higher standard deviation). We also compare this variant with a non-RFT variant of our best baseline, REAP, denoted as \textsc{REAP -- rft} to determine whether RFT is required only by our method or if it is an unavoidable by-product of compression. To this end, we observe that \pruner's \textsc{-- rft} variant outperforms \textsc{REAP -- rft} by almost $1\%$, albeit with a marginally worse cross-task standard deviation. More importantly, these results clearly demonstrate that compression induces a significant perturbation in the model's output distribution. 
RFT allows the fine-grained re-calibration of this output distribution so that model performance is recovered to an extent. Therefore, RFT is a vital part of the pruning pipeline and plays an essential role in maximizing post-compression model alignment and performance.

\endgroup

\begin{table*}[t]
\centering
% Ablations
\resizebox{\textwidth}{!}{%
\begin{tabular}{cccccc|cc}
\hline
\textbf{\# Calibration Samples} & \textbf{Generative} & \textbf{World} & \textbf{Domain} & \textbf{NLU \& NLI} & \textbf{Safety} & \textbf{Avg RP (\%)} & \textbf{Std RP (\%)} \\
& (Log-PPL $\downarrow$) & (Acc $\uparrow$) & (Acc $\uparrow$) & (Acc $\uparrow$) & (Score $\uparrow$) & ($\uparrow$) & ($\downarrow$) \\
\hline
250 & 2.18 & 64.68 & 57.30 & 73.92 &  57.97 & 98.90 & \textbf{1.90}\\
\cdashline{1-8}
50 & 2.20 & 63.95 & 56.43 & 74.00 & 57.75 & 98.22 & 2.30\\
100 & 2.19 & 64.87 & 56.58 & 74.26 & 57.76 & 98.77 & 2.20\\
500 & 2.19 & 65.28 & 56.98 & 74.03 & 57.83 & \textbf{98.93} & 2.20\\
\hline
\end{tabular}%
}
\caption{Effect of modifying the number of calibration samples used by \pruner\ to estimate inter-expert transitions. While performance seems to be near constant across all regimes, cross-task variance fluctuates by a much larger degree. All results have been obtained after RFT. Best results are in boldface.}
\label{table:calibration-data-size-comparison}
\end{table*}
\paragraph{Impact of Size of Calibration Dataset.}
We also analyze the effect of the number of calibration samples used to accumulate routing trajectories on post-compression performance and provide these results in Table~\ref{table:calibration-data-size-comparison}. We calibrate \pruner\ on 50, 100, and 500 samples and observe that in all cases, its average performance retention remains almost the same, differing by less than a percent across all variants. However, the cross-task standard deviation for each variant exhibits a much larger shift from the baseline. 
When calibrated on 50 samples, \pruner\ suffers from a relatively $21.05\%$ higher cross-task standard deviation while under the 100 and 500 sample regimes, its standard deviation increases by $15.79\%$. We hypothesize that smaller calibration sets (50 and 100 samples) do not provide \pruner\ with enough context tokens to model expert transitions holistically, forcing it to rely on the limited number of domains spanned by the tokens and optimizing for performance primarily on these domains, leading to higher cross-task variance when evaluated on a diverse set of benchmarks. On the other hand, larger calibration sets (500 samples) tend to induce a frequency-based distributional bias towards domains that are most commonly seen in the data, leading to better performance on related benchmarks, but a disproportionate loss of performance in other domains. Recognizing this behavior, we opted to utilize 250 calibration samples as this is a fair middle-ground that provides \pruner\ with enough cross-domain context to model routing behavior in a generalizable manner without biasing it towards any one particular domain, as reflected by its excellent performance and consistency.
\begin{figure}[t]
    \centering
    \begin{minipage}[t]{0.45\linewidth}
        \centering
        \includegraphics[width=\linewidth]{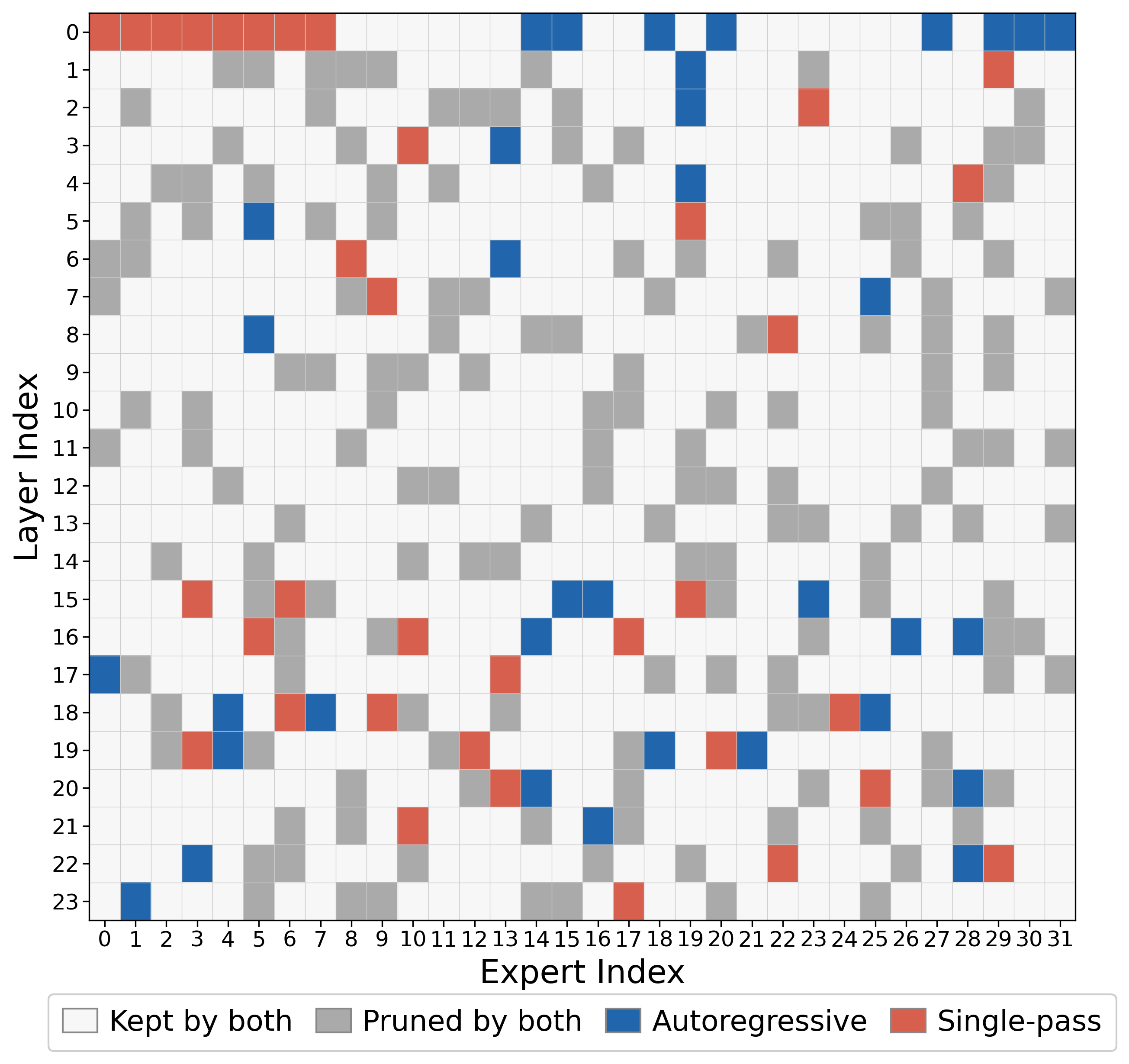}
        \captionof{figure}{Pruning decisions made by \pruner\ versus those obtained by ablating its autoregressive transition counting to naive expert activation frequency, across 24 layers of GPT-OSS-20B at 25\% compression. As evident, loss of autoregressive context leads to qualitatively different and inferior pruning decisions.}
        \label{fig:pruning-decision-comparison}
    \end{minipage}
    \hfill
    \begin{minipage}[t]{0.5\linewidth}
        \centering
        \includegraphics[width=\linewidth]{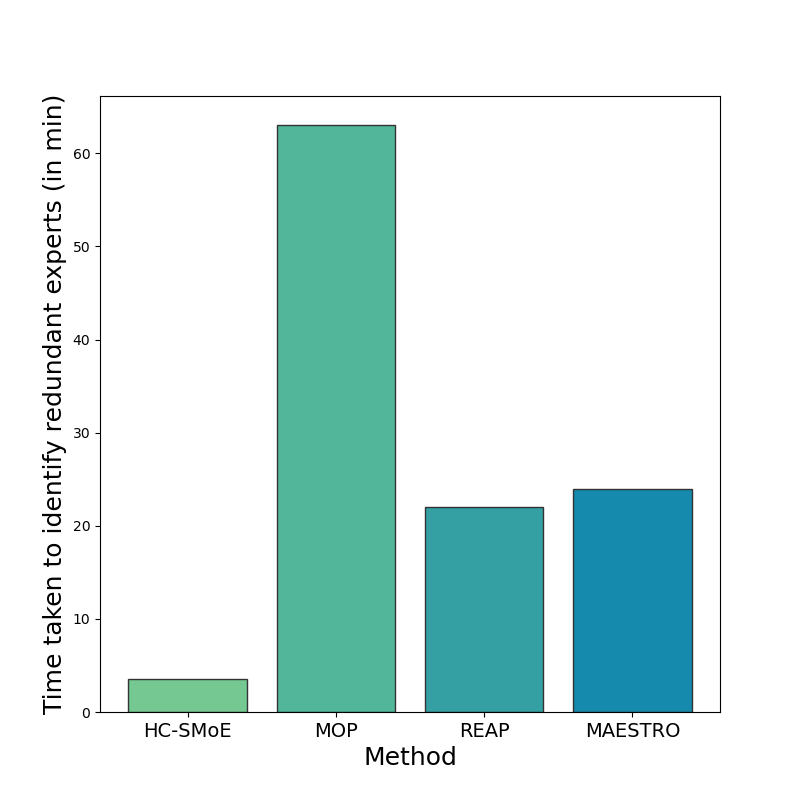}
        \captionof{figure}{Comparison of the run-times of different pruning methods on GPT-OSS-20B at a compression ratio of 25\%. Since the RFT process is method-agnostic, we omit it for this analysis.}
        \label{fig:runtime}
    \end{minipage}
\end{figure}
\section{Discussion}
\label{sec:discussion}
% \noindent\textbf{Ablation Analysis.} 
% To demonstrate the effectiveness and necessity of each component of \pruner, we conduct a thorough ablation study, the results of which are reported in Table~\ref{table:ablations}. We test four ablative variants: \textsc{--UNIFORM} (pruning experts non-uniformly from each layer), \textsc{--AUTO} (counting transitions over a single forward pass instead of over $T_{seq}$ autorgressive steps), and \textsc{--RFT} (not performing recovery fine-tuning after compression).
% Table~\ref{table:ablations} reports the performance of ablative variants of \pruner\
\paragraph{Uniform v/s Non-Uniform Pruning.}
% We explore a natural extension of \pruner\ wherein we forego the uniform pruning constraint, as detailed in Section~\ref{sec:method:selection}, and prune experts based solely on their corresponding stationary probabilities. Despite being more flexible, the latter approach suffers from a crucial drawback, \textit{i.e.}, it can lead to the over-pruning of certain layers, leading to larger irrecoverable degradations in performance. 
Figure~\ref{fig:uniform-non-uniform-comparison} demonstrates that under a non-uniform regime, \pruner\ tends to over-prune certain layers, even compressing them by almost twice as much as their uniformly pruned counterparts. For instance, layers 9, 11, 13, and 23 are pruned by 12 to 14 experts, forcing these layers to rely on a disproportionately small set of experts, 
% removes 14 out of 32 total experts from the last layer, compressing it by almost twice as much as its uniformly pruned counterpart. Layers 9, 11, and 13 undergo similar degrees of compression as well. Such an over-pruning of certain layers leads to a major performance bottleneck, forcing these layers to rely on a disproportionately small set of experts. 
leading to a performance bottleneck that outweighs the benefits of a dynamically allocated compression budget, causing the non-uniform variant to perform worse by a full percentage point on average and also exhibit higher per-task standard deviation.
% even though the non-uniform regime offers more flexibility by allocating the compression budget dynamically across layers instead of statically, it results in more unstable decisions, leading to worse and more inconsistent performance retention when compared with its uniform counterpart.  
% We analyze the differences in pruning decision and average performance under uniform and non-uniform pruning regimes
% Figure\ref{fig:uniform-non-uniform-comparison} demonstrates the differences in pruning decisions and average performance between uniform and non-uniform pruning paradigms. 
\paragraph{Influence of Autoregressive Expert Transition Counting on Pruning Decisions.}
Figure~\ref{fig:pruning-decision-comparison} demonstrates clear differences in information encoded when expert transitions are accumulated over $T_{seq} = 100$ 
autoregressive steps versus when they are computed over a single pass. The divergence is most pronounced in Layer 0, where single-pass counting prunes a near-contiguous block of low-index experts (0--7), reflecting a positional activation bias rather than any meaningful structural property of the routing graph. \pruner, by contrast, selects an entirely disjoint set in this layer, identifying experts whose low stationary probability under the autoregressive Markov chain marks them as genuinely peripheral to the model's learned routing topology. This distinction matters: early layers process token-level representations that heavily influence downstream routing decisions, so errors in pruning at this depth are likely to compound across layers. Conversely, Layers 9--14 exhibit near-total agreement between the two heuristics (grey), suggesting that a subset of experts are unambiguously redundant and recoverable by any reasonable importance metric. 
% We provide corresponding task-specific performance results in Appendix~\ref{appendix:additional-ablations}.
% Beyond these extremes, the blue cells --- experts pruned exclusively by \pruner\ --- are distributed diffusely across expert indices throughout the network, consistent with the stationary distribution surfacing globally under-utilized experts that do not cluster around any particular index or layer range. The red cells, by contrast, exhibit mild index clustering, further corroborating that frequency counting introduces a systematic low-activation bias that is an artifact of ignoring inter-expert transition structure rather than a true signal of redundancy. 
% \noindent\textbf{Support for Rapid Iteration.}
% \pruner\ has excellent fidelity for rapid iteration, a key characteristic of practical development processes. Unlike methods such as 
\paragraph{Method Run-time Comparison.}
Figure~\ref{fig:runtime} compares the time it takes for each method to compress an instance of GPT-OSS-20B by 25\%. We observe that HC-SMoE, an expert merging baseline, operates the fastest due to its rapid agglomerative clustering approach, taking roughly 3.5 minutes for the same. This excellent efficiency, however, comes at a notable cost in retained performance, as shown in Table~\ref{tab:gpt-qwen-main-results}. In contrast, MoP, a stronger baseline, takes around 63 minutes for compression, owing to its expensive expert similarity process, followed by iterative clustering. REAP completes the task in approximately 22 minutes, benefiting from its routing-based expert selection which avoids exhaustive pairwise comparisons. \pruner, our proposed method, achieves compression in around 24 minutes, a time that is comparable to REAP and around $2.5\times$ faster than MoP, while delivering superior retained performance (Table~\ref{tab:gpt-qwen-main-results}). \pruner's modest overhead over REAP is attributable to its autoregressive counting of inter-expert transitions as opposed to the single-pass importance scores that REAP operates on. Overall, \pruner\ strikes a strong balance between computational efficiency and compression quality, making it a practical choice for large-scale MoE compression.      
\section{Conclusion}
In this work, we present, \pruner, a novel structured pruning framework designed specifically for MoE LLMs. By leveraging Markov chains to model autoregressive routing decisions, \pruner\ grounds its decisions in globally relevant heuristics rather than locally-derived, single-pass activation statistics. \pruner\ outperforms state-of-the-art baselines by up to 10\% even under an extreme compression ratio of 50\%. Additionally, \pruner\ exhibits superior cross-task consistency and robustness across multiple MoE architectures.      
\section{Limitations}
We identify two main limitations with our work. First, \pruner's routing model rests on a first-order Markov assumption: the experts selected at layer $\ell+1$ depend only on those selected at layer $\ell$, not on earlier layers. While this yields a tractable and theoretically grounded stationary distribution, it may underestimate the importance of experts whose relevance is conditioned on longer routing histories; exploring higher-order chain approximations is a natural avenue for future work. Furthermore, \pruner\ currently operates at the granularity of whole experts: it either retains or discards an expert in its entirety. This binary decision may be suboptimal in settings where a moderate compression budget would be better served by a combination of expert pruning and intra-expert weight reduction, and integrating \pruner's global routing signal with width- or depth-level pruning criteria represents another promising direction.
\section{Ethical Considerations}
This work presents \pruner, a structured pruning framework for compressing 
MoE LLMs. We reflect on the ethical dimensions of this research below.

\paragraph{Usage of Existing Artifacts.}
This work builds upon publicly available model checkpoints and datasets. 
Specifically, we use GPT-OSS-20B~\citep{openai2025gptoss120bgptoss20bmodel} and Qwen3-30B~\citep{yang2025qwen3technicalreport}, accessed via their 
respective HuggingFace repositories in accordance with their published model 
cards and terms of use. For calibration and recovery fine-tuning, we use the 
SlimOrca dataset~\citep{SlimOrca}, which is released under the Apache 
2.0 license. Evaluation is conducted using the Language Model Evaluation 
Harness~\citep{lm-eval-harness}, and LoRA-based fine-tuning is performed via the PEFT library~\citep{peft}, both of which are open-source. We strictly adhere to the terms of use of each of the artifacts used by us and will release our own under the MIT License upon publication. 

\paragraph{Intended Use.} 
\pruner\ is designed to reduce the memory footprint and deployment cost of MoE models, with the goal of making them more accessible in resource-constrained settings. While we believe democratizing access to capable models is broadly beneficial, we acknowledge that lowering the barrier to deployment also lowers the barrier to misuse. Compressed models that fit on commodity hardware may be more easily deployed in adversarial contexts without institutional oversight. We encourage users of this framework to adhere to the intended use policies of the underlying model families.

\paragraph{Safety and Bias Under Compression.}
A central concern with any model compression technique is whether it disproportionately degrades safety-relevant capabilities. To this end, we explicitly evaluate all methods on Safety, Bias, and Ethics benchmarks (Winogender, TruthfulQA, and Moral Stories), which is notably absent from most prior pruning work. Our results indicate that \pruner's safety-relevant scores remain broadly stable across compression ratios, and are competitive with or superior to baselines. Nevertheless, we caution that benchmark performance on these tasks is an imperfect proxy for real-world safety behavior, and we strongly recommend that practitioners conduct targeted 
safety evaluations before deploying any compressed model in a production setting.

\paragraph{Environmental Impact.}
Model compression is, at its core, a technique for reducing the computational resources required for inference, and we therefore expect the net effect of 
this work on energy consumption to be positive. That said, the calibration and recovery fine-tuning stages of \pruner\ do incur a one-time computational cost. 

\paragraph{Broader Societal Impact.}
By enabling high-quality compression of state-of-the-art MoE models, this work contributes to a trend of increasing model accessibility. We believe this is net positive for researchers and practitioners in low-resource settings, but note that it does not address the upstream costs of training the base models, which remain substantial and are concentrated among a small number of well-resourced organizations.

\section*{Acknowledgment}
The authors acknowledge the support of the NVIDIA Academic Grant Program. T. Chakraborty acknowledges the support of the Rajiv Khemani Young Faculty Chair Professorship in Artificial Intelligence.

%%%%%%%%%%%%%%%%%%%%%%%%%%%%%%%%%%%%%%%%%%%%%%%%%

\bibliographystyle{IEEEtranN}
\small
\bibliography{custom}

%%%%%%%%%%%%%%%%%%%%%%%%%%%%%%%%%%%%%%%%%%%%%%%%%
\beginappendix
\setcounter{tocdepth}{3}

%------------RFT Table----------
\begin{table*}[!htb]
\centering
% ----------- FIRST HALF (now includes MedQA) -----------
\resizebox{\textwidth}{!}{%
\begin{tabular}{|cc|cc|ccc|cccc|}
\hline
\multirow{3}{*}{\textbf{CR}} & \multirow{3}{*}{\textbf{Method}} & \multicolumn{2}{c}{Generative} & \multicolumn{3}{|c|}{World Understanding} & \multicolumn{4}{c|}{Domain-Specific}\\
\cline{3-11}
& & \textbf{Wikitext} & \textbf{Lambada} & \textbf{PIQA} & \textbf{PROST} & \textbf{CommonsenseQA} & \textbf{OpenbookQA} & \textbf{ARC-Easy} & \textbf{ARC-Challenge} & \textbf{MedQA} \\
 & & (Log-PPL $\downarrow$) & (Log-PPL $\downarrow$) & (Acc $\uparrow$) & (Acc $\uparrow$) & (Acc $\uparrow$) & (Acc $\uparrow$) & (Acc $\uparrow$) & (Acc $\uparrow$) & (Acc $\uparrow$) \\
\hline
\multirow{1}{*}{-} & Base & 2.46 & 1.87 & 78.13 & 45.93 & 72.81 & 39.80 &  79.08 & 52.39 & 64.81\\
\cdashline{1-11}
\multirow{5}{*}{25\%} 
 & Random & 2.68 & 2.09 & 77.80 & 46.32 & 66.67 & 38.80 & 74.58 & 47.18 & 49.57\\
 & HC-SMoE & 2.85 & 2.24 & 70.78 & 35.80 & 66.99 & 36.40 & 62.79 & 43.52 & 45.56\\
 & MOP & 2.64 & 1.99 & 77.04 & 43.69 & 71.25 & 40.20 & 73.23 & 48.21 & 50.51\\
 & REAP & 2.53 & 1.89 & 77.37 & 44.44 & 70.76 & 39.60 & 74.54 & 49.57 & 58.13\\
 & \pruner\ & 2.50 & 1.86 & 78.35 & 45.03 & 70.68 & 40.00 & 78.11 & 50.60 & 60.49\\
\hline
\multirow{5}{*}{50\%} 
 & Random & 3.15 & 2.74 & 71.76 & 35.32 & 49.96 & 34.60 & 65.40 & 42.92 & 36.21\\
 & HC-SMoE & 3.27 & 2.74 & 68.50 & 36.21 & 64.46 & 32.40 & 54.97 & 34.64 & 35.04\\
 & MOP & 2.87 & 2.25 & 72.52 & 36.97 & 70.68 & 35.60 & 60.77 & 37.54 & 38.39\\
 & REAP & 2.77 & 2.16 & 73.56 & 35.40 & 66.50 & 37.40 & 62.29 & 41.47 & 37.63\\
 & \pruner\ & 2.74 & 2.06 & 77.20 & 37.90 & 66.83 & 39.00 & 69.02 & 42.49 & 51.93\\
 \hline
\end{tabular}%
}

\textit{(continued)}

% ----------- SECOND HALF (starts from BLIMP now) -----------
\resizebox{\textwidth}{!}{%
\begin{tabular}{|cc|ccccc|ccc|cc|}
\hline
\multirow{3}{*}{\textbf{CR}} & \multirow{3}{*}{\textbf{Method}} & \multicolumn{5}{c}{NLU \& NLI} & \multicolumn{3}{|c|}{Safety, Bias \& Ethics} & \multirow{3}{*}{\textbf{Avg RP (\%)}} & \multirow{3}{*}{\textbf{Std RP (\%)}}\\
\cline{3-10}
& & \textbf{BLIMP} & \textbf{BoolQ} & \textbf{Lambada} & \textbf{Winogrande} & \textbf{CoQA} & \textbf{Winogender} & \textbf{TruthfulQA} & \textbf{Moral Stories} & & \\
& & (Acc $\uparrow$) & (Acc $\uparrow$) & (Acc $\uparrow$) & (Acc $\uparrow$) & (Acc $\uparrow$) & (Acc $\uparrow$) & (Acc $\uparrow$) & (Acc $\uparrow$) & ($\uparrow$) & ($\downarrow$) \\
\hline
\multirow{1}{*}{-} & Base & 81.04 & 86.70 & 60.61 & 67.64 & 74.80 & 63.06 & 58.57 & 52.88 & - & -\\
\cdashline{1-12}
\multirow{5}{*}{25\%} 
 & Random & 80.67 & 85.47 & 57.42 & 65.11 & 73.30 & 62.92 & 57.52 & 55.53 & 95.38 & 6.00\\
 & HC-SMoE & 79.84 & 85.54 & 55.68 & 67.17 & 74.50 & 63.61 & 54.90 & 53.27 & 90.46 & 8.60\\
 & MOP & 79.66 & 84.56 & 58.55 & 67.48 & 74.00 & 61.11 & 55.86 & 52.81 & 95.63 & 5.10\\
 & REAP & 78.36 & 86.48 & 60.66 & 68.51 & 73.84 & 62.92 & 56.66 & 52.98 & 97.67 & 2.70\\
 & \pruner\ & 79.11 & 86.85 & 60.72 & 68.43 & 74.50 & 63.47 & 57.22 & 53.22 & \textbf{98.90} & \textbf{1.90}\\
\hline
\multirow{5}{*}{50\%} 
 & Random & 78.35 & 78.53 & 46.98 & 58.17 & 64.00 & 60.56 & 56.13 & 56.78 & 83.92 & 11.50\\
 & HC-SMoE & 79.21 & 82.72 & 47.41 & 64.09 & 72.40 & 60.42 & 54.94 & 53.37 & 83.71 & 12.50\\
 & MOP & 79.90 & 86.06 & 54.73 & 65.90 & 72.90 & 57.92 & 53.49 & 53.15 & 88.41 & 10.80\\
 & REAP & 77.10 & 85.02 & 56.67 & 64.48 & 72.60 & 61.11 & 55.24 & 53.52 & 89.37 & 10.40\\
 & \pruner\ & 78.15 & 84.92 & 57.37 & 65.75 & 73.00 & 61.25 & 53.52 & 53.66 & \textbf{92.59} & \textbf{6.30}\\
 \hline
\end{tabular}%
}
\caption{Detailed task-wise performance results for various pruners on GPT-OSS-20B across compression ratios of 25\% and 50\% with RFT. Best results are highlighted in boldface.}
\label{tab:gpt-oss-detailed-results}
\end{table*}

\appendix
\begin{table*}[!htb]
\centering
% ----------- FIRST HALF (now includes MedQA) -----------
\resizebox{\textwidth}{!}{%
\begin{tabular}{|cc|cc|ccc|cccc|}
\hline
\multirow{3}{*}{\textbf{CR}} & \multirow{3}{*}{\textbf{Method}} & \multicolumn{2}{c}{Generative} & \multicolumn{3}{|c|}{World Understanding} & \multicolumn{4}{c|}{Domain-Specific}\\
\cline{3-11}
& & \textbf{Wikitext} & \textbf{Lambada} & \textbf{PIQA} & \textbf{PROST} & \textbf{CommonsenseQA} & \textbf{OpenbookQA} & \textbf{ARC-Easy} & \textbf{ARC-Challenge} & \textbf{MedQA} \\
 & & (Log-PPL $\downarrow$) & (Log-PPL $\downarrow$) & (Acc $\uparrow$) & (Acc $\uparrow$) & (Acc $\uparrow$) & (Acc $\uparrow$) & (Acc $\uparrow$) & (Acc $\uparrow$) & (Acc $\uparrow$) \\
\hline
\multirow{1}{*}{-} & Base & 2.52 & 1.42 & 80.69 & 48.47 & 79.12 & 44.80 & 78.91 & 56.23 & 73.21\\
\cdashline{1-11}
\multirow{4}{*}{25\%} 
 & Random & 2.69 & 1.68 & 76.33 & 36.73 & 74.94 & 42.00 & 64.94 & 47.70 & 59.62\\
 & HC-SMoE & 2.89 & 1.81 & 70.89 & 36.71 & 71.25 & 36.00 & 64.02 & 43.77 & 42.81\\
 & REAP & 2.42 & 1.44 & 79.92 & 45.50 & 81.16 & 44.40 & 72.31 & 52.13 & 70.23\\
 & \pruner\ & 2.35 & 1.38 & 81.18 & 48.39 & 80.43 & 44.80 & 76.05 & 52.65 & 72.74\\
\hline
\multirow{4}{*}{50\%} 
 & Random & 3.22 & 2.50 & 71.44 & 27.73 & 64.78 & 33.00 & 62.67 & 42.49 & 41.71\\
 & HC-SMoE & 3.58 & 2.89 & 61.26 & 26.88 & 53.97 & 32.40 & 51.35 & 34.73 & 27.97\\
 & REAP & 2.83 & 2.07 & 74.21 & 41.52 & 74.94 & 39.80 & 70.33 & 50.00 & 54.05\\
 & \pruner\ & 2.61 & 1.46 & 79.38 & 36.85 & 81.25 & 42.60 & 72.18 & 48.55 & 44.93\\
 \hline
\end{tabular}%
}

\textit{(continued)}

% ----------- SECOND HALF (starts from BLIMP now) -----------
\resizebox{\textwidth}{!}{%
\begin{tabular}{|cc|ccccc|ccc|cc|}
\hline
\multirow{3}{*}{\textbf{CR}} & \multirow{3}{*}{\textbf{Method}} & \multicolumn{5}{c}{NLU \& NLI} & \multicolumn{3}{|c|}{Safety, Bias \& Ethics} & \multirow{3}{*}{\textbf{Avg RP (\%)}} & \multirow{3}{*}{\textbf{Std RP (\%)}}\\
\cline{3-10}
& & \textbf{BLIMP} & \textbf{BoolQ} & \textbf{Lambada} & \textbf{Winogrande} & \textbf{CoQA} & \textbf{Winogender} & \textbf{TruthfulQA} & \textbf{Moral Stories} & & \\
& & (Acc $\uparrow$) & (Acc $\uparrow$) & (Acc $\uparrow$) & (Acc $\uparrow$) & (Acc $\uparrow$) & (Acc $\uparrow$) & (Acc $\uparrow$) & (Acc $\uparrow$) & ($\uparrow$) & ($\downarrow$) \\
\hline
\multirow{1}{*}{-} & Base & 82.95 & 88.72 & 64.62 & 70.88 & 76.10 & 66.53 & 52.56 & 50.71 & - & -\\
\cdashline{1-12}
\multirow{4}{*}{25\%} 
 & Random & 81.97 & 81.50 & 63.21 & 70.80 & 75.00 & 59.31 & 51.88 & 50.80 & 91.79 & 7.10\\
 & HC-SMoE & 84.07 & 73.49 & 60.43 & 69.61 & 72.70 & 64.17 & 53.32 & 51.12 & 87.47 & 11.00\\
 & REAP & 82.36 & 85.51 & 67.26 & 71.98 & 75.80 & 57.92 & 54.25 & 50.61 & 98.13 & 4.30\\
 & \pruner\ & 82.18 & 84.65 & 67.88 & 72.06 & 75.60 & 59.44 & 52.71 & 49.69 & \textbf{99.40} & \textbf{3.50}\\
\hline
\multirow{4}{*}{50\%} 
 & Random & 76.50 & 49.82 & 50.17 & 61.80 & 67.00 & 58.47 & 49.55 & 54.51 & 78.77 & 13.50\\
 & HC-SMoE & 80.95 & 73.46 & 42.07 & 62.12 & 69.10 & 57.08 & 49.71 & 54.30 & 74.58 & 16.60\\
 & REAP & 81.26 & 81.84 & 56.59 & 69.22 & 73.50 & 56.67 & 54.95 & 51.73 & 90.25 & \textbf{7.10}\\
 & \pruner\ & 81.41 & 82.78 & 66.18 & 73.09 & 75.10 & 57.92 & 53.66 & 49.84 & \textbf{93.40} & 10.50\\
 \hline
\end{tabular}%
}
\caption{Detailed task-wise performance results for various pruners on Qwen3-30B across compression ratios of 25\% and 50\% with RFT. Best results are highlighted in boldface.}
\label{tab:qwen-detailed-results}
\end{table*}

\section{Detailed Per-Task Performance}
\label{app:detailed-results}

Tables~\ref{tab:gpt-oss-detailed-results} and~\ref{tab:qwen-detailed-results} 
report per-task performance for all methods on GPT-OSS-20B and Qwen3-30B 
respectively, across both compression ratios.

\paragraph{GPT-OSS-20B.}
At 25\% compression, \pruner\ recovers near-lossless generative quality, with 
Lambada Log-PPL ($1.86$) marginally surpassing the unpruned base ($1.87$) and 
PIQA accuracy ($78.35$) also exceeding the base ($78.13$), suggesting that 
removing low-stationary-probability experts can have a mild regularising effect 
on specific tasks. The most pronounced gains over baselines appear in 
domain-specific tasks: \pruner\ leads on MedQA by a substantial margin at both 
compression levels ($60.49$ vs. $58.13$ for REAP at 25\%; $51.93$ vs. $37.63$ 
for REAP at 50\%), indicating that salient domain-specialized experts are well-preserved by the stationary distribution heuristic. At 50\% compression, performance on NLU tasks degrades gracefully and remains competitive with or above baselines, while Safety and Ethics scores 
are broadly stable across all methods, suggesting these capabilities are 
robust to expert pruning at the ratios studied.

\paragraph{Qwen3-30B.}
The results on Qwen3-30B reinforce and strengthen the trends observed on 
GPT-OSS-20B. At 25\% compression, \pruner\ exceeds the base model on four 
tasks, Wikitext Log-PPL ($2.35$ vs. $2.52$), Lambada Log-PPL ($1.38$ vs. 
$1.42$), PIQA ($81.18$ vs. $80.69$), and CommonsenseQA ($80.43$ vs. $79.12$), pointing to a consistent regularization effect that is more pronounced in 
the larger model. Even at 50\% compression, \pruner\ retains a Log-PPL 
of $1.46$ on Lambada, within $0.04$ of the uncompressed base, a margin that no other 
method approaches. HC-SMoE degrades most severely at this compression level, 
dropping to $74.58\%$ average RP with a standard deviation of $16.60\%$, 
highlighting the instability of clustering-based approaches under aggressive 
pruning. Overall, the per-task breakdown confirms that \pruner's aggregate 
gains reported in the main results are not driven by outlier tasks but reflect 
broad, consistent performance retention across generative, knowledge, domain, 
and language understanding benchmarks.

\end{document}